\documentclass{article}

\usepackage[final,nonatbib]{neurips_2019}

\usepackage[utf8]{inputenc}
\usepackage[T1]{fontenc}
\usepackage{hyperref}
\usepackage{url}
\usepackage{booktabs}
\usepackage{amsfonts}
\usepackage{amsmath}
\usepackage{nicefrac}
\usepackage{microtype}
\usepackage{xcolor}
\usepackage{lipsum}
\usepackage{subfigure}
\usepackage{graphicx}
\usepackage{biblatex}
\usepackage{graphicx}
\usepackage{subcaption}
\usepackage{float} % in your preamble

\title{
  GPT Meets Graphs and KAN Splines: Testing Novel Frameworks on Multitask Fine-Tuned GPT-2 with LoRA \\
  \vspace{1em}
  \small{\normalfont Stanford CS224N Default Project}  % Select one and delete the other
}

\author{
  Marc Bernardino \\
  Department of Computer Science \\
  Stanford University \\
  \texttt{mrbernar@stanford.edu} \\
  \And
  Gabriel Bo \\
  Department of Computer Science \\
  Stanford University \\
  \texttt{gabebo@stanford.edu} \\
  \And
  Justin Gu \\
  Department of Computer Science \\
  Stanford University \\
  \texttt{justingu@stanford.edu}
}

\begin{document}

\maketitle
\begin{abstract} \label{abstract}
We explore the potential of integrating learnable and interpretable modules—specifically Kolmogorov-Arnold Networks (KAN) and graph-based representations—within a pre-trained GPT-2 model to enhance multi-task learning accuracy. Motivated by the recent surge in using KAN and graph attention (GAT) architectures in chain-of-thought (CoT) models and debates over their benefits compared to simpler architectures like MLPs, we begin by enhancing a standard self-attention transformer using Low-Rank Adaptation (LoRA), fine-tuning hyperparameters, and incorporating L2 regularization. This approach yields significant improvements. To further boost interpretability and richer representations, we develop two variants that attempt to improve the standard KAN and GAT: Graph LoRA and Hybrid-KAN LoRA (Learnable GPT). However, systematic evaluations reveal that neither variant outperforms the optimized LoRA-enhanced transformer, which achieves 55.249\% accuracy on the SST test set, 99.18\% on the CFIMDB dev set, and 89.9\% paraphrase detection test accuracy. On sonnet generation, we get a CHRF score of 42.097. These findings highlight that efficient parameter adaptation via LoRA remains the most effective strategy for our tasks: sentiment analysis, paraphrase detection, and sonnet generation.
\end{abstract}

\section{Key Information to include}
\underline{Mentor}: No, \underline{External Collaborators}: No, \underline{Sharing project}: No

\section{Introduction and Related Works} \label{intro}
\noindent Recent advancements in interpretable neural architectures, such as Kolmogorov-Arnold Networks (KAN) and Graph Attention Networks (GAT), have attracted interest for boosting both accuracy and interpretability in NLP \cite{liu2024kan} \cite{gat} \cite{shukla2024kan}. These models leverage B-splines on network edges and graph representations of text, respectively, which offer potential improvements to learning on tasks like sentiment analysis, paraphrase detection, and sonnet generation. However, empirical validations in multi-task settings with pretrained language models remain limited \cite{liu2024kan} \cite{young2024kan}.

\noindent To explore these claims, we integrated KAN and GAT modules into established GPT-2 architectures for multi-task benchmarks. Our approach used Low-Rank Adaptation (LoRA) for efficient fine-tuning, combined with hyperparameter optimization and L2 weight decay for robust comparisons \cite{hu2021lora}. We further enhanced these modules by developing Hybrid KAN-LoRA (Learnable GPT) and Graph-LoRA variants \cite{jiang2024hybrid} to test if these interpretable extensions could outperform conventional transformers.

\noindent Unexpectedly, our evaluations revealed that neither Hybrid KAN-LoRA nor Graph-LoRA improved performance in multi-task scenarios. Our best-performing models relied solely on conventional self-attention mechanisms enhanced by LoRA, regularization, and dropout. Quantitatively, our optimized LoRA-based transformer achieved 55.249\% accuracy on the Stanford Sentiment Treebank, 99.0\% on the CFIMDB development set, and robust paraphrase detection (90.2\% on dev, 89.9\% on test), along with competitive sonnet generation. These results suggest that for multi-task learning, the simplicity and efficiency of LoRA-enhanced transformers are preferable to the added complexity of interpretable architectures like KAN and GAT.

\noindent Recent related research has generated enthusiasm for KANs, which use spline-based adaptive activations to decompose complex functions into localized components, allowing for deep reasoning on challenging structured learning problems e.g., AIME, Physics Olympiad, and research) \cite{liu2024kan}\cite{yang2024endowinginterpretabilityneuralcognitive}. Similarly, GAT explicitly models relational structures, with Liu et al. (2024) suggesting that KANs enhance performance by enabling structured learning for difficult reasoning tasks \cite{liu2024kan}\cite{gat}. Yet, analyses such as Yu et al. (2024) question KANs' efficacy compared to simpler MLPs \cite{kanvsmlp}. This motivates our evaluation of integrating KAN and graph-based modules with LoRA, leading to Hybrid KAN-LoRA and Graph-LoRA \cite{jiang2024hybrid}. Our results, detailed in Sections \ref{results} and \ref{analysis}, underscore the need for rigorous comparisons against simpler baselines.

\section{Approach} \label{approach}

\subsection{Baseline Model} \label{base}

As noted in Section \ref{intro}, for the baseline model, we implemented GPT-2 containing byte-pair encoding (BPE) tokenization, a learnable embedding layer with position embeddings, and 12 transformer layers, similar to the original GPT-2 paper by Open-AI \cite{gpt2}. Each transformer layer utilizes masked (causal) multi-head self-attention, position-wise feed-forward networks with two linear transformations and a ReLU activation function, and dropout.

To improve the baseline model, we fine-tuned the following hyperparameters: hidden layer dropout probability and L2 regularization through weight decay \cite{l2reg} \cite{dropout} for robustness to overfitting, model size for performance improvements, and number of epochs for additional model learning.

\subsection{Low-Rank Adaptation (LoRA)} \label{lora approach}

We first extended the baseline model by implementing Low-Rank Adaptation (LoRA), which reduces the number of trainable parameters by encoding task-specific parameter increments in a low-rank representation \cite{hu2021lora}.

As described in the LoRA paper, Hu et al. (2021) the parameter increment update is expressed by the following equation:
\begin{equation}
    W = W_0 + BA  \label{LoRA}
\end{equation}

where $ W_0 \in \mathbb{R}^{d \times k} $ is a pre-trained weight matrix, and $ B \in \mathbb{R}^{d \times r} $ and $ A \in \mathbb{R}^{r \times k} $ with $r << \min(d, k)$ \cite{hu2021lora}. The initial weight matrix $W_0$ is frozen during training and thus does not receive gradient updates, while $A$ and $B$ are matrices containing tunable parameters.

Hence, the final forward pass can be expressed as $h = W_0 x + BA_x$, where the product $BA$ represents the learned low-rank adaptation of the weights. Thus, LoRA theoretically increases training efficiency and allows for the fine-tuning of larger models with more parameters. By integrating LoRA into our transformer model, we aimed to explore its effectiveness in three downstream language tasks.

\subsection{Kolmogorov-Arnold Networks (KANs)} \label{kan approach}

Our primary research goal was to implement Kolmogorov-Arnold Network (KAN) layers as an alternative to traditional Multi-Layer Perceptrons (MLPs) for our transformer architecture. While MLPs have fixed activation functions at each layer’s nodes, KANs utilize learnable activation functions along edges, replacing linear weights with univariate functions parameterized as splines.

As described by Liu et al. (2024), KANs are derived from the Kolmogorov-Arnold representation theorem, which states that any multivariate continuous function $f: [0, 1]^n \rightarrow \mathbb{R}$ can be expressed as a sum of continuous univariate functions \cite{liu2024kan}:

\begin{equation} f(\textbf{x}) = f(x_1, \cdots, x_n) = \sum_{q=1}^{2n+1} \Phi_q \left( \sum_{p=1}^n \phi_{q,p} (x_p) \right) \label{kan equation 1}\end{equation}

where $\Phi_q: \mathbb{R} \rightarrow \mathbb{R}$ and $\phi_{q,p}: [0, 1] \rightarrow \mathbb{R}$.

\begin{figure}
  \centering
  % First subfigure
  \begin{subfigure}
    \centering
    \includegraphics[width=0.45\textwidth]{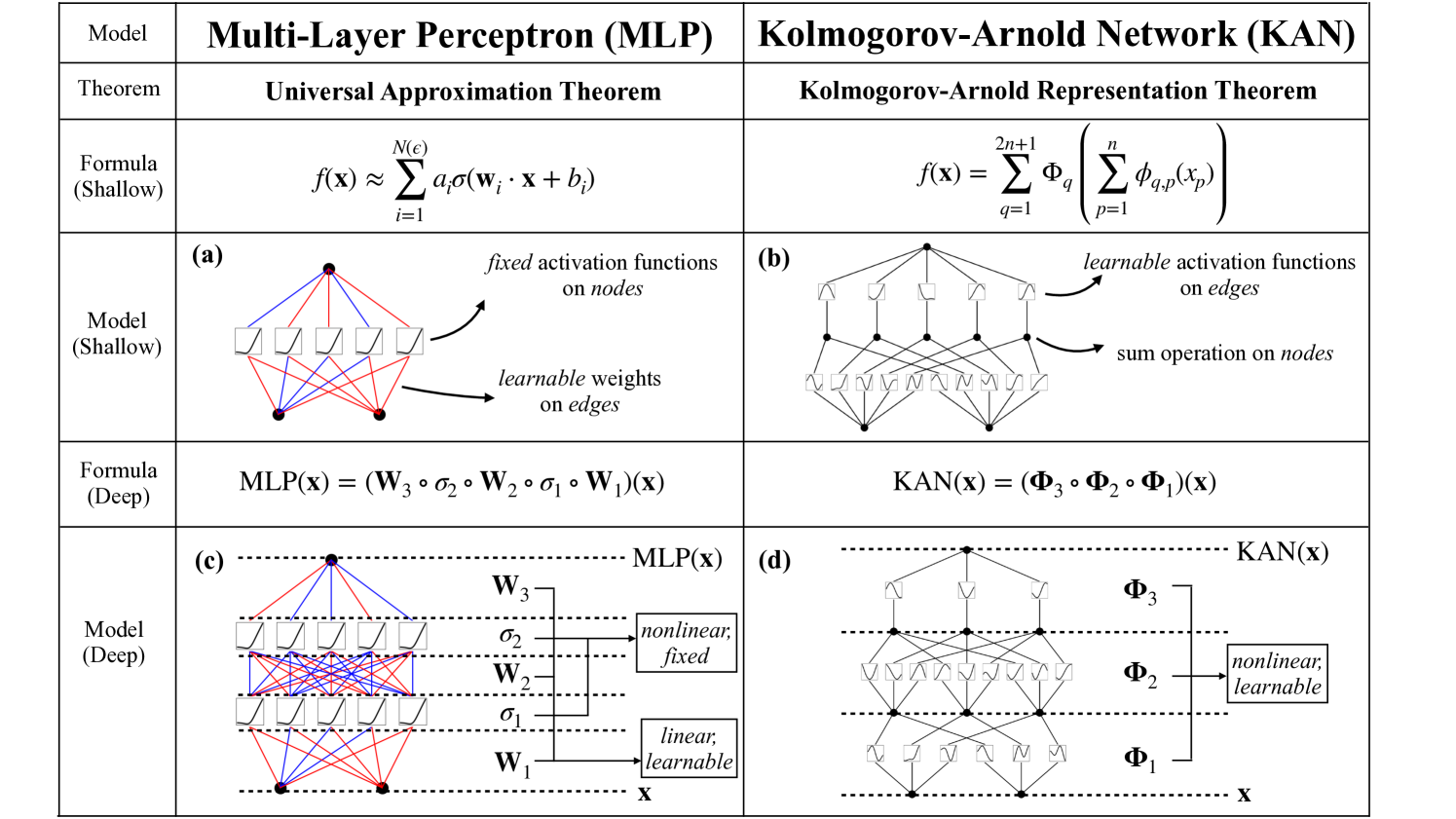}
  \end{subfigure}
  \hspace{1em}
  % Second subfigure
  \begin{subfigure}
    \centering
    \includegraphics[width=0.45\textwidth]{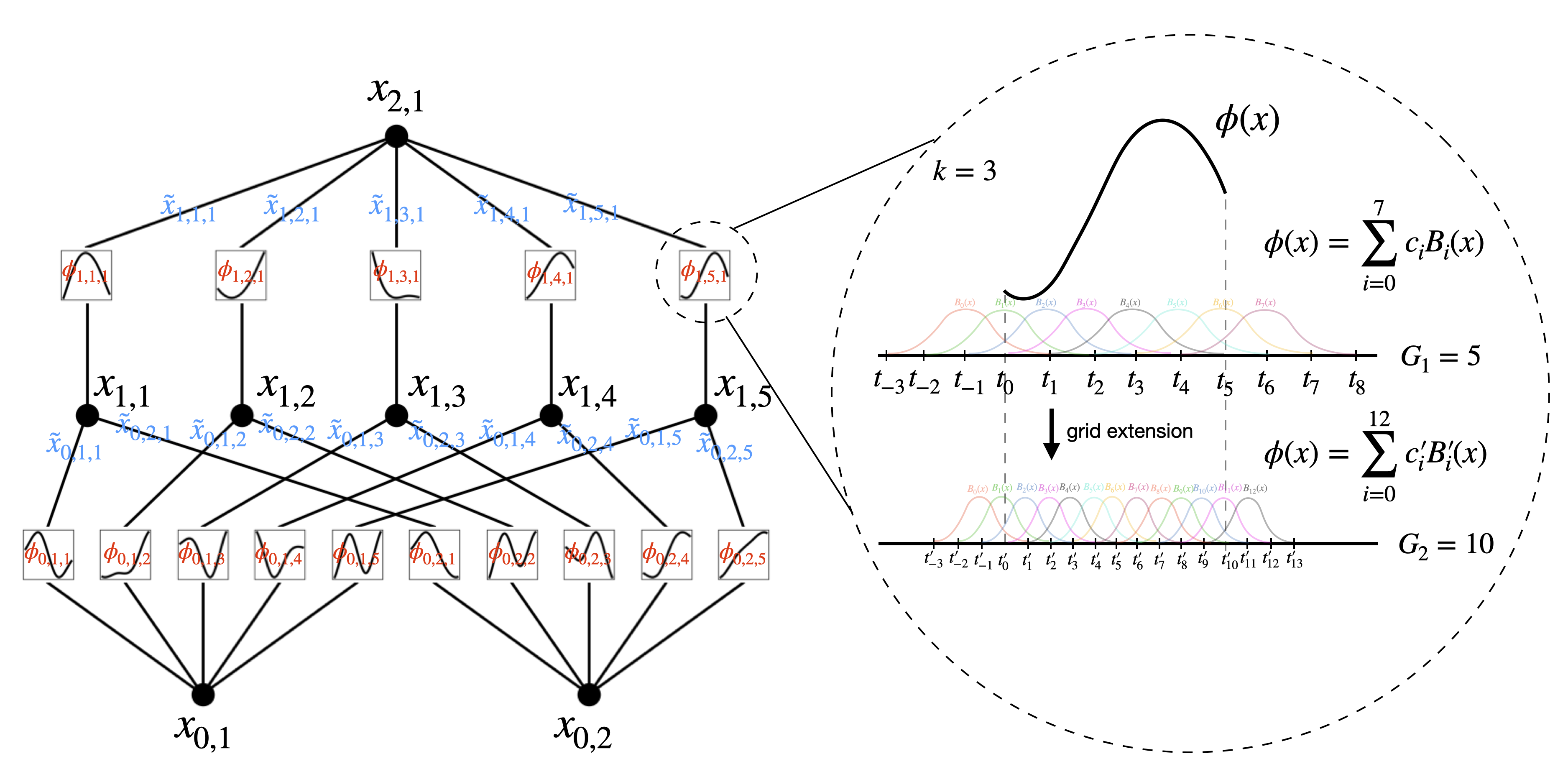}
  \end{subfigure}
  \caption{KAN compared to standard MLP architecture from Liu et al. (2024)}
  \label{fig:combined_kan}
\end{figure}

KANs approximate this representation by approximating each $\phi$ with learnable spline functions $\sigma_{KAN}(x)$ at each node and grid-based representations. As displayed in Figure \ref{fig:combined_kan}, each spline-based transformation is implemented as a linear combination of B-splines, such that:

\begin{equation} \sigma_{\text{KAN}}(x) = \sum_k s_k B_k(x) \label{kan equation 2} \end{equation}

where $s_k$ are learnable spline coefficients.

Liu et al. (2024) highlight KANs' benefits in mathematics and physics-related tasks which commonly have smooth functions \cite{liu2024kan}. Thus, we explore whether such learnable activation functions could enhance text-based NLP models by replacing MLPs in transformer feedforward layers (this decision can be better understood from Figure \ref{fig:combined_kan}).

\subsection{Graph Attention Networks (GATs)} \label{gat approach}

Recent research suggests that graph attention architectures could be used in text classification and generation tasks. Graph neural networks (GNNs) represent texts in a graph-based structure, where nodes resemble words or tokens, and edges resemble semantic relationships. Wang et al. (2024) classify over twenty GNN text models into two types: corpus-level GNNs, which represent an entire corpus as a graph, and document-level GNNs, which construct graphs per document \cite{Wang_2024} \cite{koncelkedziorski2022textgenerationknowledgegraphs}.

We implement a document-level GNN approach, applying a multi-head Graph Attention Network (GAT) layer to propagate contextual information between nodes. Our implementation is derived from the original GAT paper published by Veličković et al. (2018) and modeled after Figure \ref{fig: gat} \cite{gat}.

The multi-head attention mechanism is characterized by the concatenation of K attention heads. Given an input graph represented by a node feature matrix H and an adjacency matrix A, the multi-head GAT operation that updates the node representation is defined as follows:

\begin{equation}
\vec{h}'_i = \bigg\|_{k=1}^{K} \sigma \left( \sum_{j \in \mathcal{N}_i} \alpha_{ij}^k \textbf{W}^k \vec{h}_i \right) \label{gat attention head equa}
\end{equation}

where $\|$ represents concatenation. $\textbf{W}$ is a learnable weight matrix, $\mathcal{N}_i$ represents the neighbors of node $i$, and $\alpha_{ij}^k$ is the $k$-th head’s attention coefficient, with an individual attention head coefficient computed as:

\begin{equation}
\alpha_{ij}^{(k)} = \text{softmax}(\text{LeakReLU}(\textbf{a}^\top [\textbf{W}h_i \| \textbf{W}h_j])) \label{attention heads}
\end{equation}

Beyond text classification, we aim to extend this graph attention layer to text generation as well, building off of the successes of Koncel-Kedziorski et al. (2022) \cite{koncelkedziorski2022textgenerationknowledgegraphs}. We also aim to explore the interaction of graph attention with our other implementations, such as examining if parameter-efficient fine-tuning methods like LoRA allow for computationally efficient graph attention methods for text modeling.

\subsection{Extending LoRA with KANs and GATs} \label{extensions of lora}

We extend our model by combining LoRA with KANs and GATs, aiming to leverage LoRA’s efficient fine-tuning with the expressiveness of KANs and structured learning of GATs.

\begin{figure}[h]
    \centering
    \includegraphics[width=0.75\linewidth]{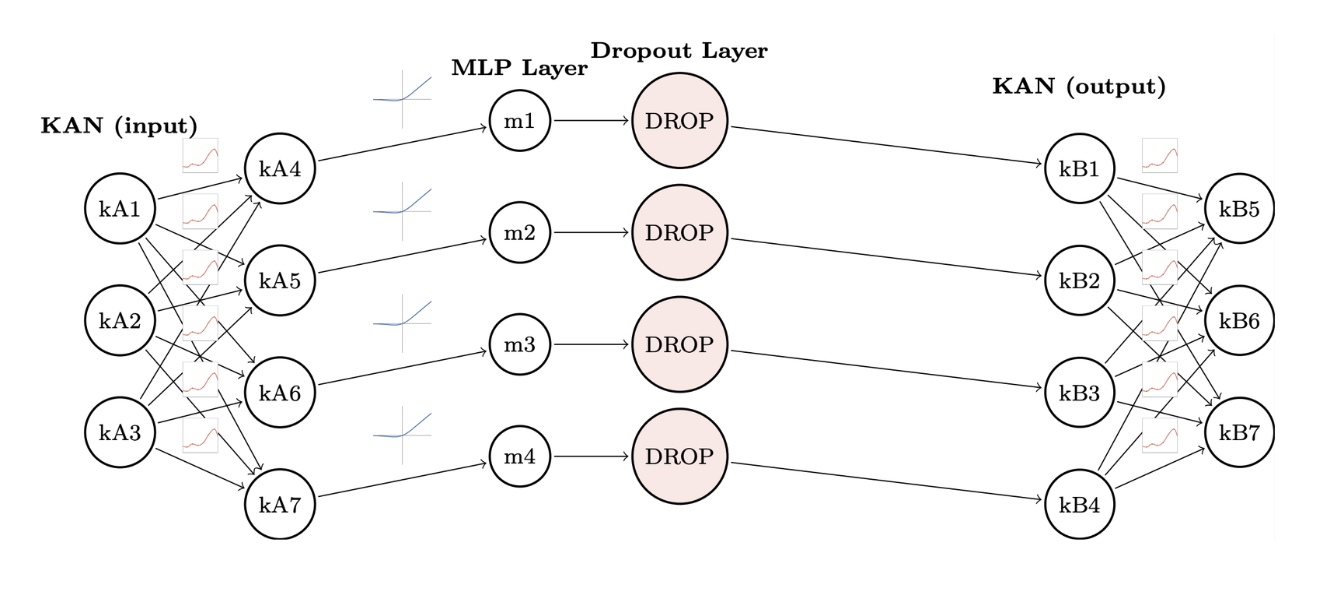}
    \caption{Hybrid KAN Layer Implementation}
    \label{fig:hybrid kan}
\end{figure}

\paragraph{\textbf{Hybrid KAN-LoRA (Learnable GPT)}} Building on Jiang et al. (2024)'s Hybrid KAN framework, we integrated LoRA into KAN to enable parameter-efficient fine-tuning (see Section \ref{experiments}) \cite{jiang2024hybrid}. As shown in Figure \ref{fig:hybrid kan}, LoRA is applied to the base path of KANLinear layers in the transformer's feedforward network, while preserving the original spline path computations. This hybrid approach fine-tunes only the linear transformations in the base path, maintaining the flexibility of B-spline activations:

\begin{equation}
h = \left( W_{\text{base}} + \frac{\alpha}{r}BA \right) f_{\text{base}}(x) + W_{\text{spline}} \sigma_{\text{KAN}}(x)
\end{equation}

\begin{figure}
  \centering
  % First subfigure
  \begin{subfigure}
    \centering
    \includegraphics[width=0.45\textwidth]{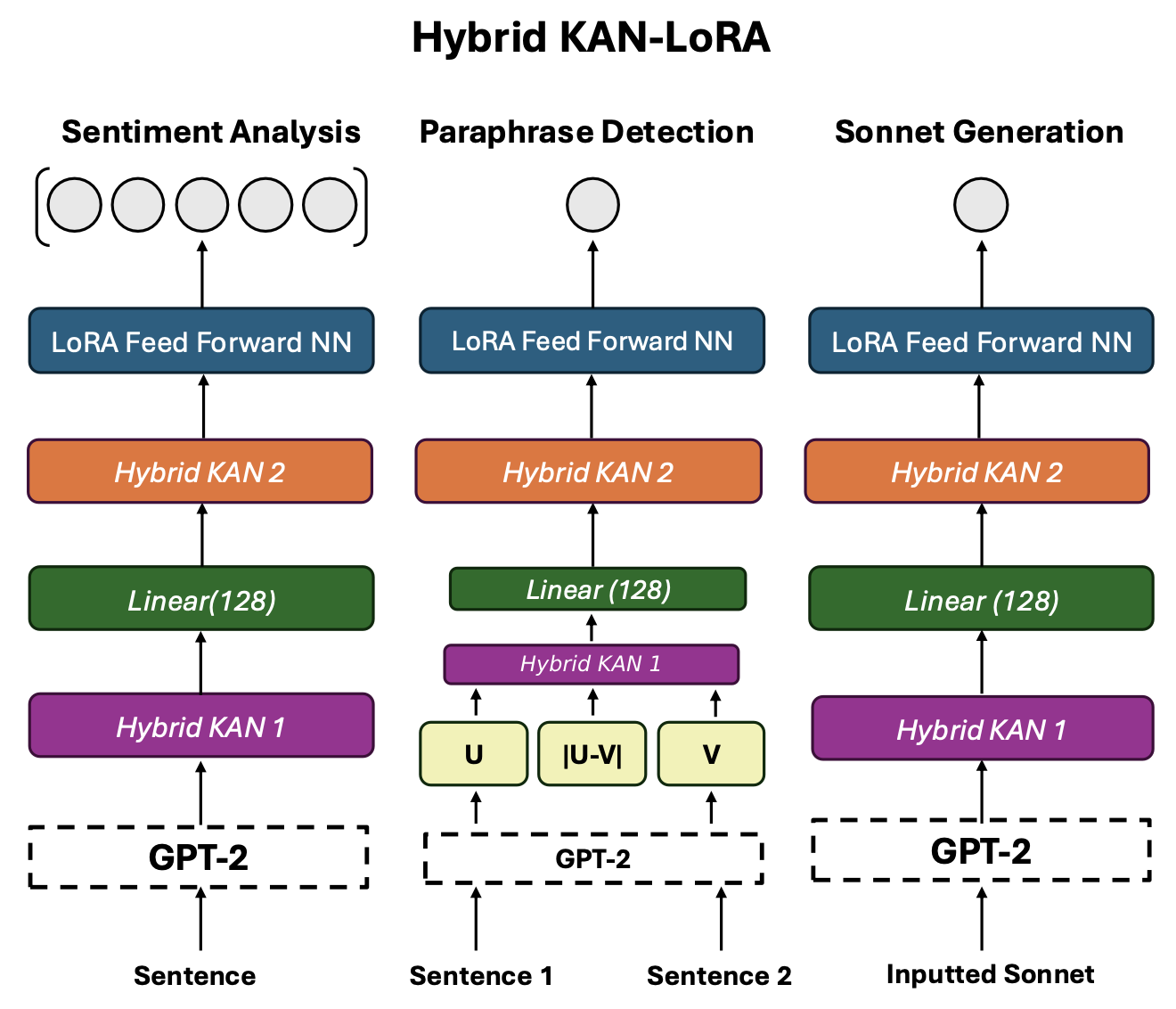}
  \end{subfigure}
  \hspace{1em}
  % Second subfigure
  \begin{subfigure}
    \centering
    \includegraphics[width=0.45\textwidth]{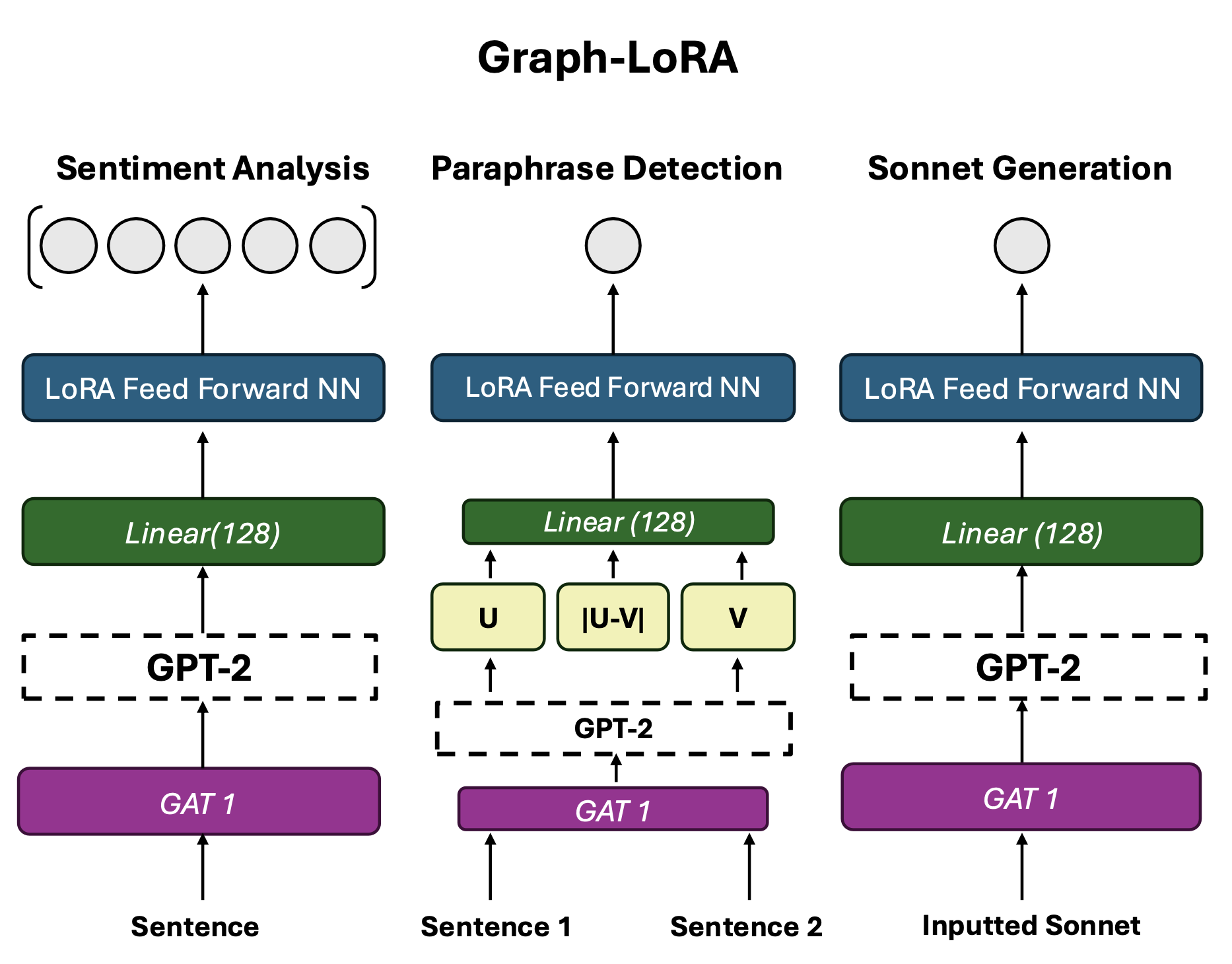}
  \end{subfigure}
  \caption{The Hybrid KAN-LoRA \& The Graph-LoRA Architecture (using Figure \ref{fig:hybrid kan}).}
  \label{fig:model arch}
\end{figure}

Since LoRA focuses on a low-rank update to only the base path weights, the KAN's spline path preserves its unique flexibility through unmodified B-spline activations\footnote{We utilize a modified version of Liu et al. (2024)'s source code \cite{liu2024kan} to speed up KAN training, titled An Efficient Implementation of Kolmogorov-Arnold Network by the GitHub user Blealtan \cite{efficient_kan}.} Coupling Figure \ref{fig:hybrid kan}'s representation of our Hybrid KAN with the LoRA implementation, we get our extended Hybrid KAN-LoRA model shown in Figure \ref{fig:model arch}.

\paragraph{\textbf{Graph-LoRA}}In our Graph-LoRA implementation, applying LoRA to GAT weight matrices reduces trainable parameters while preserving the ability to dynamically model relationships in text graphs. Given multi-head GAT operations described above, we apply LoRA to the learnable weight matrix $\textbf{W}$ such that:

\begin{equation} \textbf{W}^k = \textbf{W}_0^k + B^k A^k. \end{equation}
\cite{hu2021lora}
This implementation seeks to retain the dynamic attention weight learning while benefiting from LoRA’s parameter reduction to support efficient fine-tuning of GAT layers. The model maintains the ability to utilize pre-trained GAT weights while fine-tuning through low-rank parameter updates. \footnote{We inherent most of the GAT implementation from Koncel-Kedziorski (2022), Wang et al. (2024), and Veličković et al. (2018), with the LoRA paper source code.}

The proposed KAN-LoRA and Graph-LoRA extensions, shown in Figure  \ref{fig:model arch}, represent new approaches to integrating parameter-efficient fine-tuning with expressive activation function learning and graph-based modeling, respectively. We evaluate these interpretable model implementations on performance and parameter-efficiency with regard to downstream language modeling tasks.

\section{Experiments} \label{experiments}

\subsection{Data}
Table \ref{table: dataset} lists the datasets used: the Stanford Sentiment Treebank (SST) contains 11,855 sentences (215,154 phrases) labeled on a five-point scale \cite{sst}; CFIMDB includes 2,434 movie reviews with binary labels; Quora Question Pairs comprises 400,000 question pairs for binary paraphrase classification \cite{quora}; and the Sonnet database features 154 Shakespearean sonnets (14 lines, ABAB CDCD EFEF GG rhyme scheme).

\begin{table}[h]
    \centering
    \begin{tabular}{l l l r}
        \toprule
        \textbf{Dataset} & \textbf{Task} & \textbf{Type} & \textbf{Examples} \\
        \midrule
        Stanford Sentiment Treebank (SST) & Sentiment Analysis & Classification & 11,855 \\
        CFIMDB & Sentiment Analysis & Classification & 2,434 \\
        Quora Question Pairs & Paraphrase Detection & Classification & 400,000 \\
        Shakespearean Sonnets & Text Generation & Generation & 154 \\
        \bottomrule
    \end{tabular}
    \caption{Training and Development Datasets}
    \label{table: dataset}
\end{table}

\subsection{Evaluation method}
The two sentiment benchmarks, SST and CFIMDB, are evaluated on classification accuracy, similar to those used by Yang et al. (2024) \cite{yang2024endowinginterpretabilityneuralcognitive}. Paraphrase detection uses cloze-style binary classification accuracy, and sonnet generation is evaluated with CHaRacter-level F-score (CHRF), comparing text similarity via character n-grams \cite{huggingface2024chrf}.

\subsection{Experimental details}
In the initial step before extending GPT‑2, we ran a baseline sanity check using a simple transformer with feed-forward and MLP layers. This naive test failed to match the optimal LoRA scores on all tasks. We optimized hyperparameters: learning rate ($1e^{-5}$), dropout (0.5), and weight decay (0.2). For sentiment analysis and sonnet generation, 10–15 epochs were optimal, while for paraphrase detection, the best performance occurred at 3–5 epochs due to long compute times.

\paragraph{\textbf{Pre-training with Efficient Fine-tuning}}
In our experiments, we adapted the baseline NLP model for multiple tasks by incorporating LoRA-based low-rank adapters. The key LoRA hyperparameters---(\texttt{r}, \(\alpha\), \texttt{dropout})---were tuned to balance model expressiveness with training efficiency:

\begin{enumerate}
    \item \textbf{Low-Rank Dimension} \(\boldsymbol{(r)}\)

        The integer \(r\) controls the rank of the adapter matrices \(\texttt{lora\_A}\) and \(\texttt{lora\_B}\); higher \(r\) increases capacity but also parameters. We tested values (e.g., \(r \in \{4, 8, 16, \dots\}\)) to capture task-specific variations without overfitting, and found that \(r=32\) yielded the best performance. This is noticed from the Figure \ref{fig: lora} in the Appendix.

    \item \textbf{Scaling Factor} \(\boldsymbol{(\alpha)}\)

    As mentioned in Section \ref{lora approach}, the scaling factor \(\alpha\) (\texttt{lora\_alpha}) amplifies the low-rank update in:
\[
\text{output} = F.\text{linear}(x,W_0,b_0) + \left(\frac{\alpha}{r}\right) F.\text{linear}(F.\text{linear}(x,\texttt{lora\_A}), \texttt{lora\_B}).
\]
A moderate \(\alpha\) lets the adapters adjust without overwhelming the frozen base. We found that \(\alpha=64\) yielded the best performance, consistent with Hayou et al. (2024) \cite{lora+}.
\end{enumerate}
Eventually the final LoRA kept the same hyperparameters but also trained on a model size of 1280 dimension parameters-- a much larger model than the standard hugging face GPT-2 \cite{gpt2}.
\paragraph{\textbf{Graph-LoRA}}
The Graph-LoRA architecture described in Section \ref{gat approach} integrates LoRA into graph-based transformer models, efficiently adjusting weights via low-rank decompositions to reduce computational demands and enhance performance. 

Our experiments used optimized hyperparameters—learning rate: \(1 \times 10^{-5}\), dropout: 0.5, and weight decay: 0.2—for sentiment analysis, paraphrase detection, and sonnet generation. We employed multi-head graph attention mechanisms to boost relational learning. In our final experiment, we varied the number of attention heads \(K\) to assess performance changes. In Equation (\ref{attention heads}), the index \((k)\) in \(\alpha_{ij}^{(k)}\) denotes each head capturing unique relational patterns. We observed performance improvements up to an optimal \(K\), beyond which computational overhead increased without additional gains.

\paragraph{\textbf{Hybrid KAN-LoRA (Learnable GPT)}}
Using similar hyperparameters, the Hybrid KAN-LoRA model in Figure \ref{fig:model arch} used a learning rate of \(1 \times 10^{-5}\), dropout of 0.5, and weight decay of 0.2. We further refined the combination of LoRA fine-tuning with KAN through experiments:
\begin{itemize}
    \item \textbf{Scaling Factors.} Both LoRA (via \(\alpha\) or \(\texttt{lora\_alpha}\)) and KANs (via \(\texttt{scale\_base}\) and \(\texttt{scale\_spline}\)) include scaling parameters. Systematic variation of these factors showed that LoRA applied to the base weight only slightly influenced the final output relative to the base components.
\end{itemize}

\subsection{Results} \label{results}

\begin{table}[h]
    \centering
    \begin{tabular}{l c c c c}
        \toprule
        \textbf{Model Type} & \textbf{SST Acc} & \textbf{CFIMDB Acc} & \textbf{Para Acc} & \textbf{Sonnet CHRF}\\
        \midrule
        Baseline & 0.506 & 0.971 & 0.883 & 41.097 \\
        Baseline (GPT2-Medium) & 0.513 & 0.976 & -- & 41.397 \\
        Baseline (GPT2-Large) & 0.514 & 0.985 & 0.898 & 42.259\\
        LoRA & 0.510 & 0.98 & 0.885 & 42.096 \\
        LoRA (GPT2-Medium) & 0.512 & 0.983 & -- & 42.195 \\
        \textbf{LoRA (GPT2-Large)} & \textbf{0.516} & \textbf{0.992} & \textbf{0.902} & \textbf{42.403}\\
        KAN & 0.454 & -- & -- & -- \\
        KAN-LoRA & 0.458 & 0.723 & 0.795 & 34.433\\
        Graph & 0.443 & -- & -- & --\\
        Graph-LoRA & 0.446 & 0.704 & 0.812 & 35.904 \\
        \bottomrule
    \end{tabular}
    \caption{Development Set Performance Across Model Variants}
    \label{fig: results}
\end{table}

LoRA on the GPT2-Large model (1280-dim) achieved the strongest performance across all development metrics—SST Acc (0.516), CFIMDB Acc (0.992), Para Acc (0.902), and Sonnet CHRF (42.403). Extending this model to the paraphrase detection and sonnet generation test datasets yielded 0.899 Para Test Acc and 42.097 Sonnet Test CHRF. LoRA fine-tuning generally improved scores while reducing trainable parameters and increasing training efficiency; larger models yielded better performance, and combining them with LoRA further boosted results.

However, contrary to our expectations, our KAN, KAN-LoRA, Graph, and Graph-LoRA implementations significantly underperformed compared to both the baseline and LoRA models. In Section \ref{analysis}, we examine potential reasons for these unexpected results.

% \subsubsection{Base Results}
% \begin{minipage}{0.3\textwidth}
%     \includegraphics[scale=0.06]{sonnet2.png}
% \end{minipage}%
% \hfill
% \begin{minipage}{0.6\textwidth}
%     Here is the text describing the Base Results. You can write any explanation or details about the results here. (DATA NOT DONE)
% \end{minipage}

% \subsubsection{Paraphrase Results}
% \begin{minipage}{0.3\textwidth}
%     \includegraphics[scale=0.06]{sonnet2.png}
% \end{minipage}%
% \hfill
% \begin{minipage}{0.6\textwidth}
%     Here is the text describing the Paraphrase Results. Add relevant information or analysis for this section. (DATA NOT DONE)
% \end{minipage}

% \subsubsection{Sonnet Results}
% \begin{minipage}{0.3\textwidth}
%     \includegraphics[scale=0.06]{sonnet2.png}
% \end{minipage}%
% \hfill
% \begin{minipage}{0.6\textwidth}
%     Here is the text describing the Sonnet Results. Provide insights or observations about this subsection.
% \end{minipage}

% Report the quantitative results that you have found so far. Use a table or plot to compare results and compare against baselines.
% \begin{itemize}
%         \item If you're a default project team, you should \textcolor{red}{\textbf{report the scores you obtained on the test leaderboards for paraphrase detection and sonnet generation}}. You can also report dev set results if you like. Also mention the results after implementing your extensions.
%         \item Comment on your quantitative results. Are they what you expected? Better than you expected? Worse than you expected? Why do you think that is? What does that tell you about your approach?
% \end{itemize}

\section{Analysis} \label{analysis}
\subsection{Suboptimal Nature of \textbf{Graph-LoRA}} \label{graph-lora sucks}
\paragraph{\textbf{Key Takeaways}} 
Graph LoRA integrates low-rank adaptation into graph neural network layers to capture graph-specific relationships. However, experimental results indicate that this added complexity does not yield proportional benefits; Saparov et al. (2024) note that larger transformer models learn basic search poorly \cite{saparov}. The extra parameters and modifications result in higher computational overhead without a performance boost—echoing findings that a standard Transformer with LoRA-augmented MLP layers is simpler and more efficient.

Moreover, building a graph data structure for every transformer pass incurs significant cost: training for paraphrase detection took nearly 18 hours and 14 steps for SST training on an A100 GPU using the smallest GPT-2 model, as shown in Figure \ref{fig: sst}, \ref{fig:results}.

\subsection{Suboptimal Nature of \text{Hybrid KAN-LoRA (Learnable GPT)}} \label{kan-lora sucks}
\paragraph{\textbf{Discussion}} According to “KAN or MLP: A Fairer Comparison” by Yu et al. (2024), KAN outperforms MLP only in symbolic formula tasks \cite{kanvsmlp}. In contrast, MLP layers consistently yield better results on standard machine learning, vision, NLP, and audio tasks. Thus, even when applying LoRA, KAN’s B-spline approach lacks intrinsic benefits for most real-world tasks, adding complexity without proportional performance gains over optimized MLP layers.

\begin{figure}
  \centering
  % First subfigure
  \begin{subfigure}
    \centering
    \includegraphics[width=0.4\textwidth]{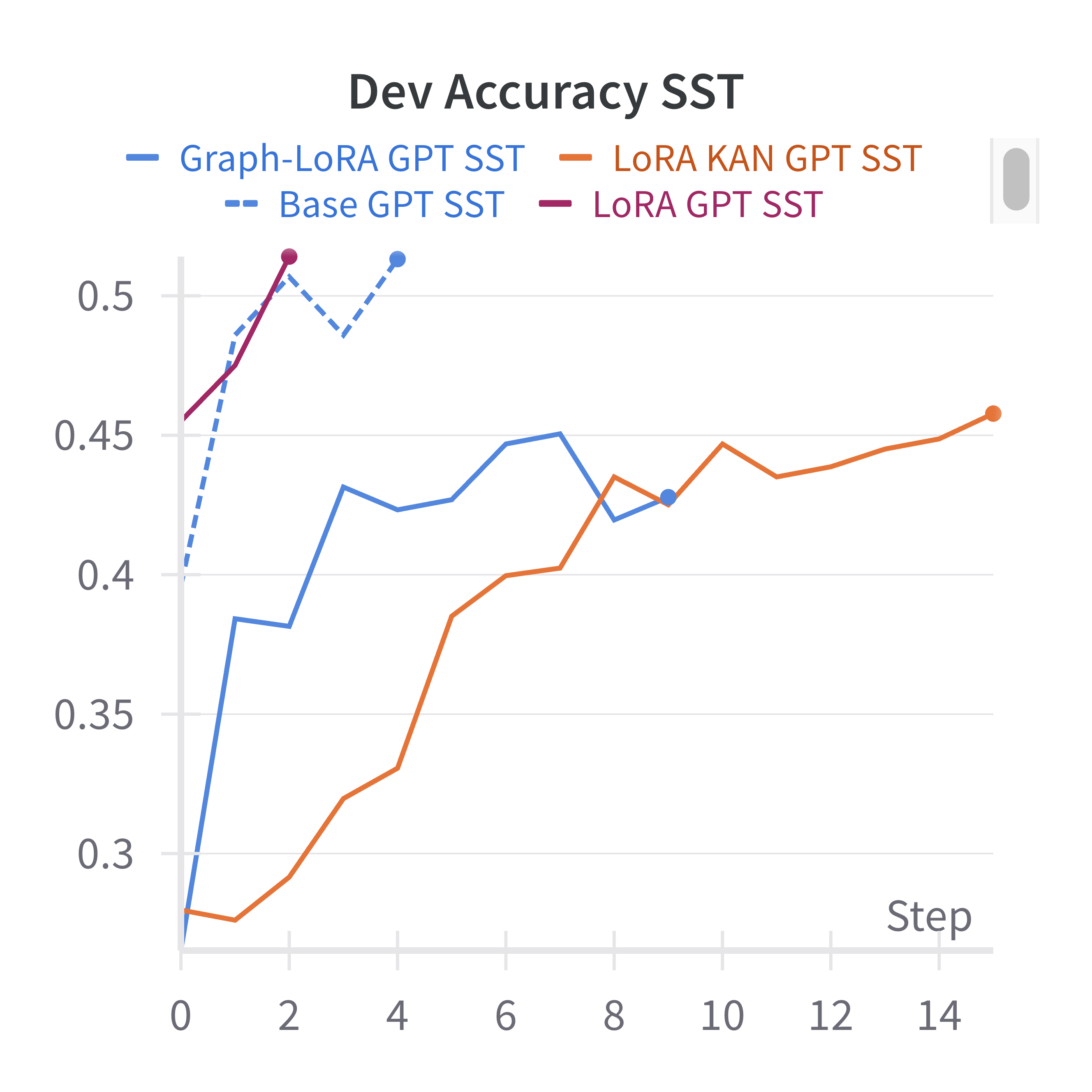}
  \end{subfigure}
  \hspace{1em}
  % Second subfigure
  \begin{subfigure}
    \centering
    \includegraphics[width=0.4\textwidth]{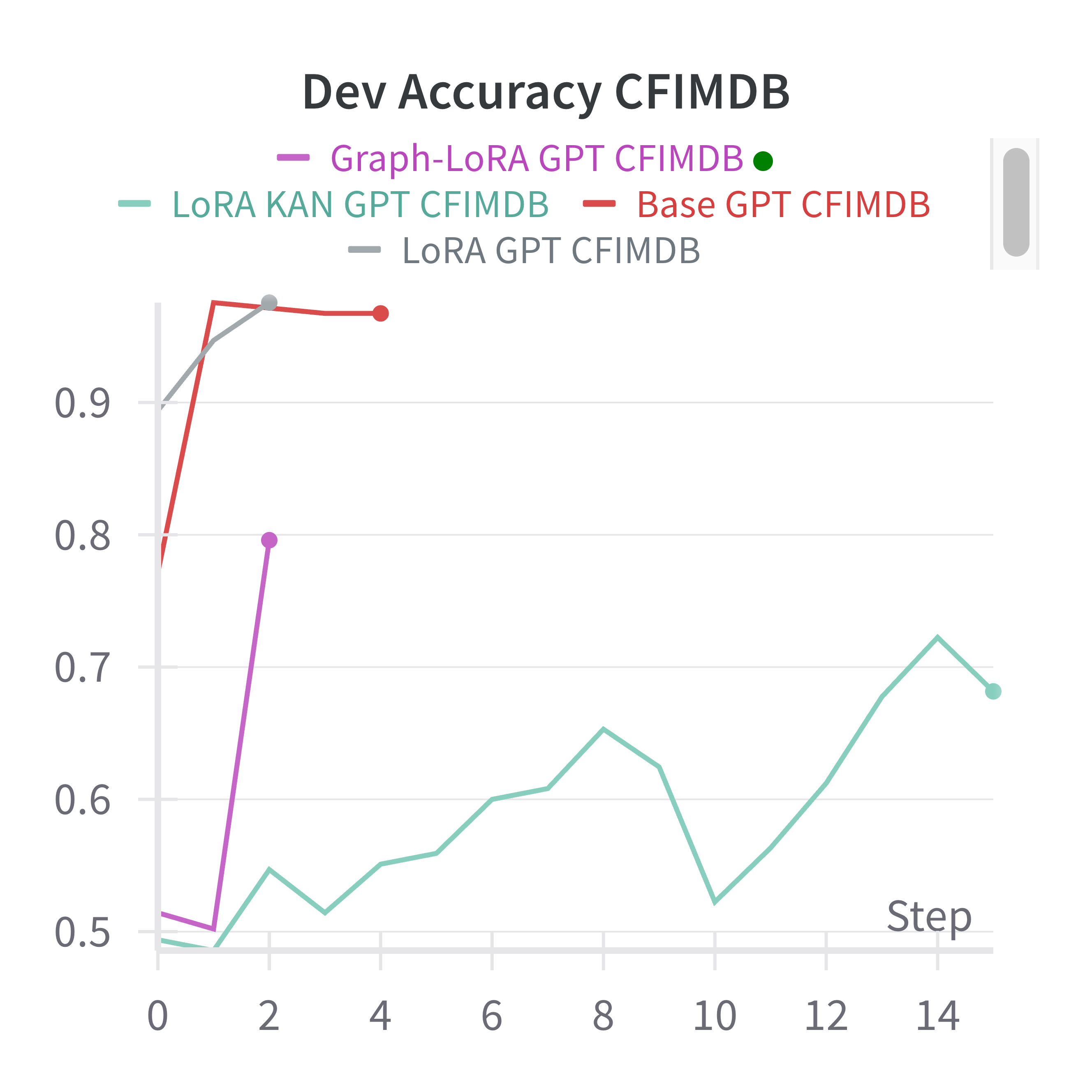}
  \end{subfigure}
  \caption{Sentiment Analysis Scores on SST and CFIMDB using architectures in Section  \ref{experiments}. Note that early stopping was used for some models to maximize performance.}
  \label{fig: sst}
\end{figure}

\paragraph{\textbf{Key Takeaways}} KAN shows promise on complex problems—resulting in a smaller performance gap between our Learnable GPT and the baseline on challenging tasks (e.g., sonnet generation) compared to simpler ones such as sentiment analysis (Figure \ref{fig: sst}). However, while LoRA’s parameter-efficient fine-tuning preserves a frozen base model, KAN’s weaker memorization can hurt its performance. We see this in Figure \ref{fig:results} when Learnable GPT takes longer to learn in the early epochs. Modern Transformers, built around optimized MLP blocks, integrate LoRA more seamlessly with their attention and MLP layers.

\subsection{Strength of LoRA}
LoRA outperformed Learnable GPT and Graph-LoRA on all multi-task metrics. Our main metrics (error analysis and performance) drew inspiration from Jiang et al. (2024) and Liu et al. (2024)’s work where they also observed the effectiveness of KAN models \cite{jiang2024hybrid}\cite{liu2024kan}.

LoRA’s strength in multitask settings comes from combining weight decay and dropout on low-rank update parameters. As shown in Equation (\ref{LoRA}), LoRA adjusts a frozen weight $W_{0}$ by adding low-rank factors $BA$. Weight decay keeps these factors small to prevent overfitting, while dropout promotes generalization across tasks. Regularizing only $B$ and $A$ lets LoRA adapt effectively without destabilizing the core, boosting accuracy, F1 scores, and other metrics across diverse tasks.

Moreover, although LoRA outperformed the baseline, the Learnable GPT and Graph-LoRA implementations were too computationally expensive to train on larger GPT-2 models. They were limited to 768 parameters compared to LoRA’s 1280, explaining the results in Sections \ref{results}, \ref{graph-lora sucks}, and \ref{kan-lora sucks}.

\begin{figure}
  \centering
  % First subfigure
  \begin{subfigure}
    \centering
    \includegraphics[width=0.4\textwidth]{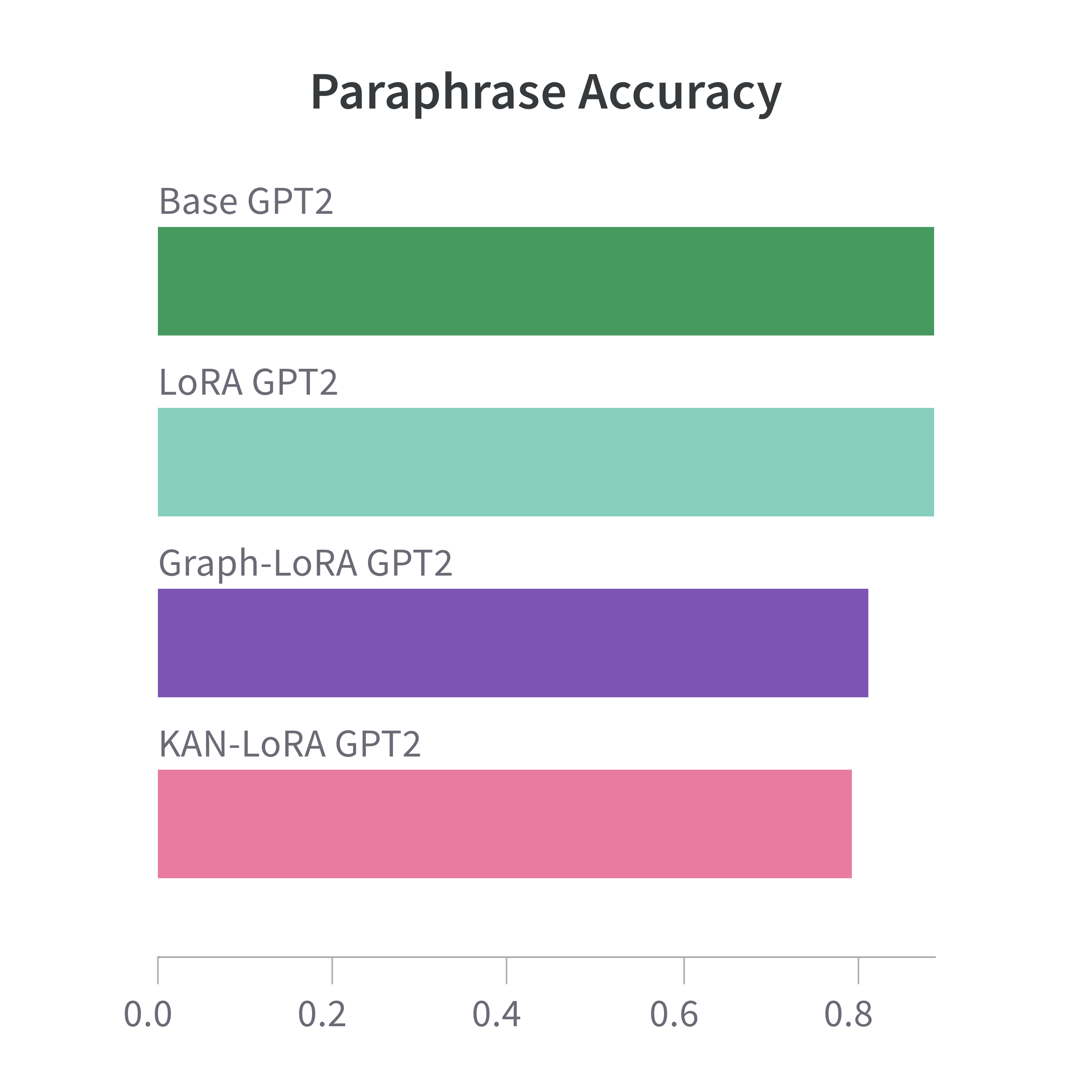}
  \end{subfigure}
  \hspace{1em}
  % Second subfigure
  \begin{subfigure}
    \centering
    \includegraphics[width=0.4\textwidth]{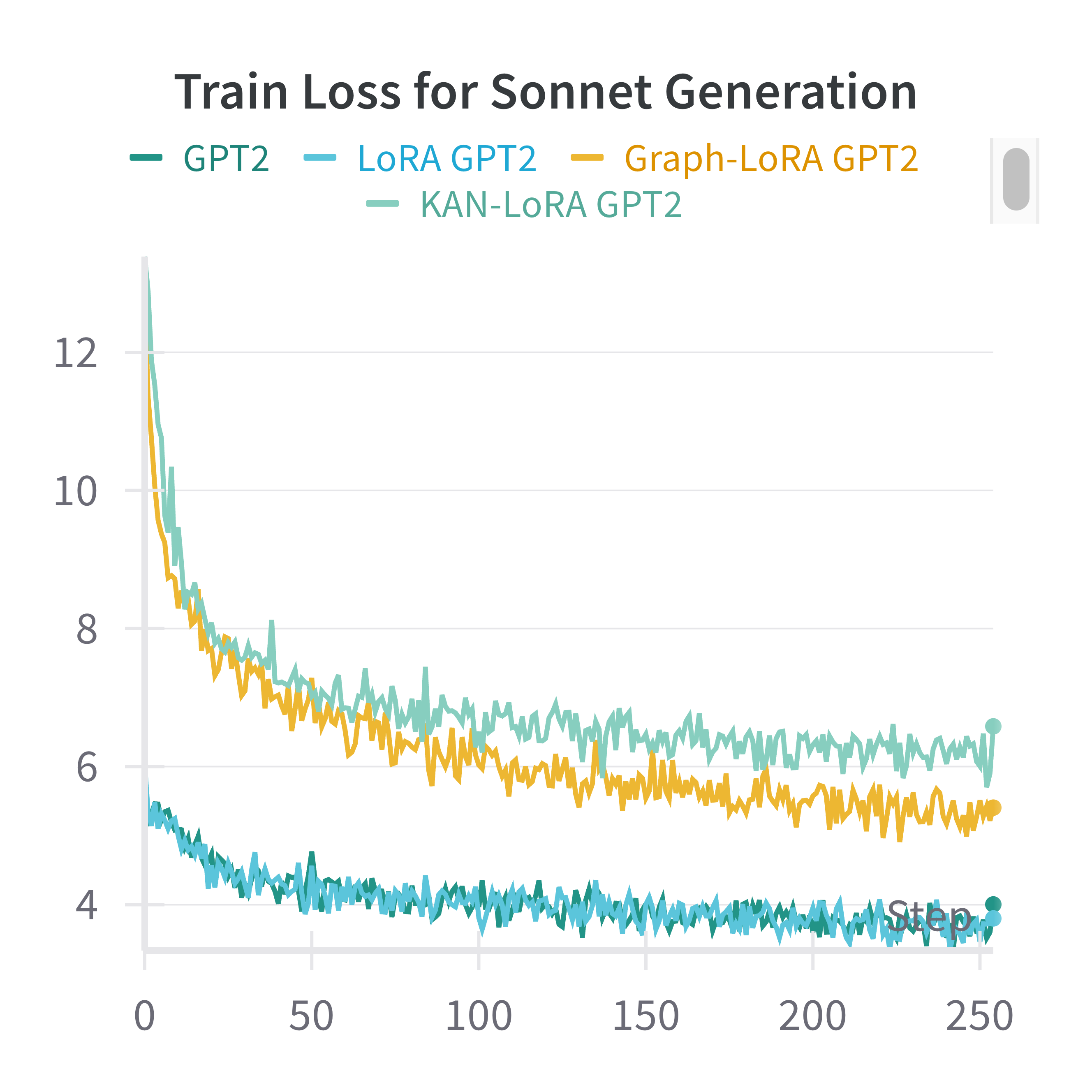}
  \end{subfigure}
  \caption{Paraphrase and Sonnet Train Loss Scores using architectures in Section \ref{experiments}.}
  \label{fig:results}
\end{figure}

Figure \ref{fig:results} shows that LoRA GPT-2 outperforms Graph-LoRA and KAN-LoRA in Sonnet Generation. Its simpler low-rank adapters retain GPT-2’s strong memorization ability, while the extra overhead and sensitivity of graph-based or spline-based modules hamper training convergence, resulting in higher loss and poorer initial test scores.
\section{Conclusion}
In this paper, we integrated Kolmogorov-Arnold Networks (KAN) and Graph Attention Networks (GAT) with Low-Rank Adaptation (LoRA) to improve multi-task learning in pretrained GPT-2 models. Despite their appeal for interpretability and structured learning, experiments showed that neither KAN-LoRA nor Graph-LoRA outperformed standard LoRA-enhanced transformers. Our optimized LoRA model consistently delivered top results in tasks such as sentiment analysis (55.249\%), paraphrase detection (89.9\%), and sonnet generation (CHRF=42.097). These findings highlight that simple, parameter-efficient approaches like LoRA are preferable to more complex architectures and stress the need for rigorous empirical validation against strong, simpler baselines.

\section*{Team contributions}
\noindent \textbf{Marc}: Developed the base model, implemented KAN and KAN-LoRA, set up Wandb logging, created figures, conducted research, and edited the paper.

\textbf{Gabriel}: Implemented LoRA with KAN and GAT, trained models on GPU, assisted with the base model, performed research, and wrote the report.

\textbf{Justin}: Built the graph attention architecture, supported the base model, managed GPU training, wrote the approach section, and conducted research.

% \bibliographystyle{unsrt}
% \bibliography{references}

\printbibliography
\appendix

\section{Appendix}
\subsection{Graph Attention}
\begin{figure}[H]
    \centering
    \includegraphics[width=0.7\linewidth]{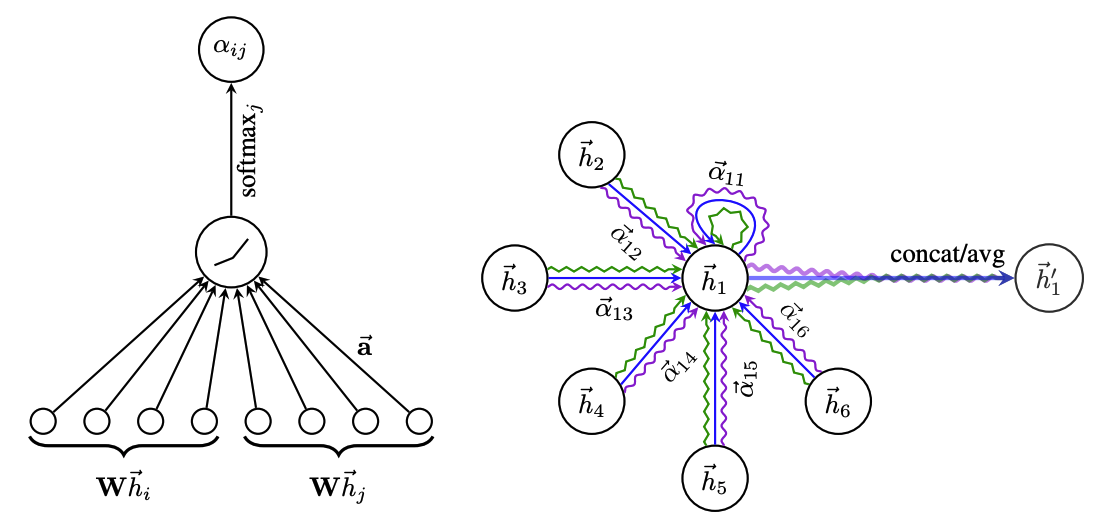}
    \caption{Left: Example of the graph node we implement in Section \ref{gat approach} and Right: the multi-head attention we apply to the transformer.}
    \label{fig: gat}
\end{figure}
In Figure \ref{fig: gat}, we utilize the architecture similar to Veličković et al. (2018)'s paper \cite{gat}. 

\subsection{LoRA}
\begin{figure}[H]
    \centering
    \includegraphics[width=0.5\linewidth]{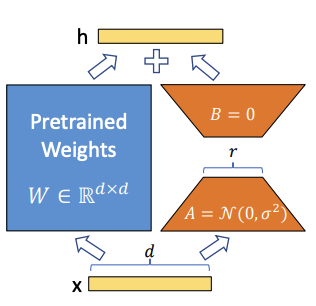}
    \caption{This is the LoRA head that represents the reduced dimensionality into a lower rank, used in Section \ref{lora approach}.}
    \label{fig: lora}
\end{figure}

We utilize this knowledge of LoRA in Hu et al. (2021) to implement our standard LoRA model but also the additional LoRA extensions with the Learnable GPT with KAN and graph attention in Graph-LoRA (found in Section \ref{extensions of lora} \cite{hu2021lora}.

\end{document}